\documentclass[10pt, a4paper]{article}

\usepackage[final]{lrec2026}
\usepackage{booktabs}
\usepackage{multirow}
\usepackage{graphicx}
\usepackage{xcolor}
\usepackage{amsmath}

\title{L2D-Clinical: Learning to Defer for Adaptive Model Selection in Clinical Text Classification}

\name{Rishik Kondadadi, John E. Ortega} 

\address{University of Minnesota, Northeastern University \\
         Minneapolis, Boston, USA  \\
         konda052@umn.edu, j.ortega@northeastern.edu\\}

\abstract{
Clinical text classification requires choosing between specialized fine-tuned models (BERT variants) and general-purpose large language models (LLMs), yet neither dominates across all instances. We introduce Learning to Defer for clinical text (L2D-Clinical), a framework that learns when a BERT classifier should defer to an LLM based on uncertainty signals and text characteristics. Unlike prior L2D work that defers to human experts assumed universally superior, our approach enables adaptive deferral---improving accuracy when the LLM complements BERT. We evaluate on two English clinical tasks: (1) ADE detection (ADE Corpus V2), where BioBERT (F1=0.911) outperforms the LLM (F1=0.765), and (2) treatment outcome classification (MIMIC-IV with multi-LLM consensus ground truth), where GPT-5-nano (F1=0.967) outperforms ClinicalBERT (F1=0.887). On ADE, L2D-Clinical achieves F1=0.928 (+1.7 points over BERT) by selectively deferring 7\% of instances where the LLM's high recall compensates for BERT's misses. On MIMIC, L2D-Clinical achieves F1=0.980 (+9.3 points over BERT) by deferring only 16.8\% of cases to the LLM. The key insight is that L2D-Clinical learns to selectively leverage LLM strengths while minimizing API costs.
\\ \newline }

\begin{document}

\maketitleabstract

\section{Introduction}

Clinical text classification is fundamental to healthcare NLP applications, from adverse drug event (ADE) detection for pharmacovigilance \citep{henry2020n2c2,karimi2015text} to treatment outcome extraction for clinical decision support \citep{xu2022survey}. The choice between specialized fine-tuned models (e.g., BioBERT \citep{lee2020biobert}, ClinicalBERT \citep{alsentzer2019publicly}) and general-purpose large language models (LLMs) is often task-dependent, with neither approach universally superior.

Domain-adapted BERT models \citep{lee2020biobert,alsentzer2019publicly} achieve strong performance when fine-tuned on specific tasks, benefiting from in-domain pretraining. LLMs offer complementary capabilities through broad training and few-shot learning, but their cost, latency, and variable performance across tasks can be limiting factors.

We propose \textbf{Learning to Defer for clinical text} (L2D-Clinical), which trains a model to predict when BERT is likely to make errors and should defer to an LLM. Unlike fixed confidence thresholds, L2D-Clinical learns from multiple signals including prediction uncertainty, text characteristics, and domain-specific features.

This setting differs fundamentally from prior L2D work, which defers to human experts assumed \textit{universally superior} when consulted. In our framework, the LLM ``expert'' may be \textit{worse overall} than BERT (F1=0.765 vs. 0.911 on ADE). Yet L2D-Clinical still improves accuracy by learning to identify the \textit{specific instances} where the weaker-overall model excels. This requires learning fine-grained complementarity rather than simply deferring when uncertain.

To validate this approach, we evaluate on two clinical NLP tasks: (1) \textbf{ADE Detection} using the ADE Corpus V2~\citep{gurulingappa2012development}, where BioBERT outperforms the LLM but selective deferral still helps, and (2) \textbf{Treatment Outcome Classification} using MIMIC-IV discharge summaries~\citep{johnson2023mimic},\footnote{\url{https://physionet.org/content/mimiciv/}} where GPT-5-nano outperforms ClinicalBERT and L2D-Clinical learns optimal deferral patterns.

\noindent These pairings reflect the distinct strengths of each model type: domain-adapted BERT models (BioBERT, ClinicalBERT) excel at pattern matching through specialized pretraining, while LLMs leverage broad world knowledge for contextual reasoning. To illustrate the complementarity we aim to exploit, consider a clinical sentence: \textit{``Patient developed rash after starting amoxicillin.''} BioBERT, trained on biomedical text, correctly identifies the ADE relationship through learned lexical patterns. However, for \textit{``The medication was well tolerated with no significant issues noted during the hospital stay,''} BioBERT's uncertainty is high because the implicit negation and hedging language fall outside its learned patterns. In such cases, deferring to an LLM that better captures semantic nuance improves accuracy. L2D-Clinical learns to make this routing decision automatically, combining BERT's prediction confidence with text features to identify when deferral is warranted.

While the individual components (BERT classifiers, LLM prompting, logistic regression) are well-established, our contribution lies in their novel combination: applying the L2D framework to AI-to-AI deferral in clinical NLP, where the ``expert'' is not universally superior---a setting not studied in prior work. Our contributions are:
\begin{enumerate}
    \item \textbf{AI-to-AI deferral}: Unlike prior L2D work that defers to human experts, we study deferral between AI systems (BERT $\rightarrow$ LLM), where the ``expert'' is not universally superior. For example, our LLM achieves only F1=0.765 on ADE detection vs. BERT's 0.911, yet selective deferral still improves overall performance.
    \item \textbf{Adaptive deferral behavior}: We demonstrate that L2D-Clinical adapts to different model dynamics---improving accuracy when deferral helps (ADE: +1.7 F1 points, MIMIC: +9.3 F1 points) while minimizing LLM usage (7\% and 16.8\% deferral rates respectively).
    \item \textbf{Consensus ground truth for MIMIC}: We introduce a multi-LLM consensus labeling methodology (GPT-5.2 + Claude 4.5 agreement) that produces higher-quality ground truth than single-model annotation, with 83.8\% agreement rate yielding 2,782 reliable labels. Human expert validation on a 5\% subset confirmed 100\% label accuracy.
    \item \textbf{Clinical notes at scale}: Experiments on real MIMIC-IV discharge summaries (279 test samples) alongside the ADE benchmark (500 samples), demonstrating applicability to actual EHR data.
    \item \textbf{Interpretable patterns}: Analysis reveals task-specific features (e.g., causal language, outcome keywords) that predict when each model excels, supporting clinical validation.
\end{enumerate}

\section{Related Work}

Our work draws on three research threads: clinical text classification methods, the learning to defer framework, and hybrid NLP systems that combine multiple models for efficient inference.

\subsection{Clinical Text Classification}

The extraction of ADEs from clinical text has been extensively studied through shared tasks and benchmark datasets. The n2c2 2018 shared task \citep{henry2020n2c2} established benchmarks for medication and ADE extraction from EHRs, with top systems achieving F1 scores around 0.94 for medication extraction but lower scores for ADE relation extraction. The ADE Corpus V2 \citep{gurulingappa2012development} provides 23,516 sentences from PubMed case reports annotated for drug-ADE relationships, enabling sentence-level classification research. \citet{karimi2015text} survey text mining techniques for ADE detection, highlighting the challenges of implicit relationships and domain-specific vocabulary.

Treatment outcome extraction from clinical notes is an emerging area with applications in drug repurposing and pharmacovigilance \citep{xu2022survey}. While structured outcome data exists in resources like clinical trial registries, extracting treatment effectiveness from narrative discharge summaries remains challenging due to implicit language, hedging, and context dependence.

Recent systematic reviews \citep{ullah2025llmhealthcare} document rapid adoption of LLMs for clinical text classification, though domain-specific BERT models remain competitive on many benchmarks. \citet{zhang2025bert} systematically compared BERT and LLM approaches across six clinical and biomedical datasets, finding that ``BERT-like models excel in pattern-driven tasks, while LLMs dominate those requiring deep semantics.'' They propose TaMAS, a task-level selection heuristic based on dataset characteristics. Our work extends this insight: rather than selecting \textit{one} model per task, we learn to select per-instance, achieving gains even when overall model rankings are clear. \citet{agrawal2022large} demonstrate that LLMs can serve as effective few-shot clinical information extractors, though with variable performance across entity types.

\subsection{Learning to Defer}

The observation that different models excel on different instances motivates the learning to defer framework, where a model learns when to defer decisions to an alternative system. \citet{madras2018predict} introduced the framework in the context of fairness, showing that selective deferral can improve both accuracy and demographic parity. \citet{mozannar2020consistent} established theoretical foundations for consistent L2D training, proving that certain surrogate losses yield Bayes-optimal deferral policies. \citet{mozannar2023teaching} provide a comprehensive survey of L2D methods, categorizing approaches by their deferral mechanisms and theoretical guarantees. \citet{raghu2019algorithmic} analyze the algorithmic automation problem, examining how to optimally allocate tasks between humans and machines.

In the medical domain, \citet{charusaie2022sample} proposed Learning to Defer with Uncertainty (LDU) for clinical diagnosis, achieving F1=0.96 on pleural effusion detection while deferring 36\% of cases to radiologists. Their work assumes the human expert is \textit{always more reliable when deferred to}, a key difference from our setting. \citet{varshney2022investigating} investigate selective prediction across clinical and general NLP tasks, finding that calibrated uncertainty estimates are crucial for effective abstention.

Our work differs from prior L2D research in three key ways. First, we study \textbf{AI-to-AI deferral} where the ``expert'' (LLM) is not universally superior; on ADE detection, our LLM achieves only F1=0.765 versus BERT's 0.911. Second, our evaluation demonstrates that L2D improves accuracy regardless of which individual model is stronger, validating the approach in both directions. Third, we target \textbf{text classification} rather than image-based diagnosis, where the complementary strengths of pattern-matching (BERT) versus semantic reasoning (LLM) create different deferral dynamics than human-AI collaboration.

\subsection{Hybrid NLP Systems}

Our approach is also related to stacked generalization \citep{ting1997stacking}, which trains a meta-learner to combine base model outputs. However, L2D-Clinical differs in that it routes to one model rather than blending predictions, and its deferral decision is interpretable. \citet{schwartz2020right} propose instance-level model selection based on complexity, showing that matching model capacity to instance difficulty improves efficiency. \citet{swayamdipta2020cartography} introduce dataset cartography, mapping training dynamics to identify easy, ambiguous, and hard instances---a complementary perspective to our uncertainty-based deferral. Our work operationalizes similar intuitions in a clinical setting, learning per-instance routing between architecturally different models.

Beyond the L2D framework, recent work addresses efficient inference through model cascading and routing. FrugalGPT \citep{chen2023frugalgpt} cascades multiple LLMs, using smaller models first and escalating to larger ones when confidence is low. They report up to 98\% cost reduction with comparable accuracy on general NLP benchmarks. Our approach differs: (1) we combine \textit{architecturally different} models (BERT vs. LLM) rather than LLM variants, (2) we use learned deferral rather than confidence cascading, and (3) we evaluate on clinical text where domain adaptation matters. Our 83--93\% LLM cost reduction with F1 improvements demonstrates L2D-Clinical's effectiveness for specialized domains.

\section{Method}

Building on these foundations, we develop a lightweight deferral model that routes inputs between BERT and the LLM. The workflow proceeds as follows: (1) BERT classifies the input text and produces softmax probabilities; (2) the deferral model examines these probabilities along with text characteristics to predict whether BERT is likely wrong; (3) if the predicted error probability exceeds a threshold, we query the LLM instead. This routing decision is learned from data, specifically by training on cases where BERT made errors on a validation set. The key insight is that BERT's uncertainty signals, combined with task-specific text features, reliably predict when deferral will help.

\subsection{Problem Formulation}

Let $x$ be an input text and $y \in \mathcal{Y}$ be the label (binary for ADE detection, 3-class for treatment outcomes). We have access to two classifiers: $f_B(x)$, a domain-adapted BERT classifier (BioBERT for ADE, ClinicalBERT for treatment outcomes), and $f_L(x)$, an LLM with few-shot prompting.

Our goal is to learn a deferral function $d(x) \in [0, 1]$ that outputs the probability of deferring to the LLM. Note that $d$ takes features derived from both the input text $x$ and BERT's output probabilities $f_B(x)$; we write $d(x)$ for brevity. Given threshold $\tau$, the final prediction is:
\begin{equation}
\hat{y} = \begin{cases}
f_L(x) & \text{if } d(x) \geq \tau \\
f_B(x) & \text{otherwise}
\end{cases}
\end{equation}

\noindent To illustrate, consider two sentences at inference time. For \textit{``Symptoms improved after switching to lisinopril,''} BERT predicts ``no ADE'' with high confidence; the deferral model outputs $d(x) = 0.12 < \tau$, so we use BERT's prediction. For \textit{``Patient reported feeling dizzy, possibly related to the new medication,''} BERT is uncertain; the deferral model outputs $d(x) = 0.73 \geq \tau$, triggering deferral to the LLM, which correctly identifies the potential ADE.

\subsection{Deferral Feature Engineering}

The deferral model needs informative features to predict when BERT will fail. We draw on two intuitions: (1) BERT's own uncertainty (reflected in its softmax distribution) signals difficult cases, and (2) certain text characteristics correlate with model-specific strengths. For instance, BERT handles explicit causal language well (``caused by,'' ``due to''), while LLMs better capture implicit relationships and hedged language.

We engineer features that capture when BERT might benefit from LLM assistance:

\paragraph{Uncertainty Features} We derive three features from BERT's softmax probability distribution $p$: (1) \textit{confidence}, defined as $\max_i p_i$, representing the model's certainty in its top prediction; (2) \textit{entropy}, computed as $H(p) = -\sum_i p_i \log p_i$, which captures overall uncertainty across all classes; and (3) \textit{margin}, the difference between the top two class probabilities, indicating decision boundary proximity.

\paragraph{Text Features} We include task-specific linguistic signals that correlate with model-specific strengths. Length features capture text complexity via $\log(1 + \text{len})$ for both character and word counts. For ADE detection, we include binary indicators for causal language (``induced,'' ``caused by,'' ``due to,'' ``after,'' ``following''), severity terms (``severe,'' ``fatal''), and common ADE terminology (``toxicity,'' ``reaction,'' ``syndrome''). For treatment outcomes, we include indicators for outcome-related terms (``discontinued,'' ``improved,'' ``intolerance''). These 18 features (5 uncertainty + 13 text) form the input to the deferral model.

\subsection{Deferral Model Training}

We train the deferral model to predict when BERT is wrong. Let $e(x) = \mathbf{1}[f_B(x) \neq y]$ indicate whether BERT erred on instance $x$. We minimize binary cross-entropy:
\begin{equation}
d^* = \arg\min_d \sum_x \mathcal{L}_{\text{BCE}}(e(x), d(x))
\end{equation}
where $\mathcal{L}_{\text{BCE}}(e, d) = -e \log d - (1-e) \log(1-d)$ is binary cross-entropy.

We use logistic regression with balanced class weights to handle the low error rate (BERT is usually correct). The threshold $\tau$ is tuned on validation data to maximize the combined system's F1 score. To prevent overfitting, we employ cross-validation: 5-fold for ADE, train/val split for MIMIC-IV.

\subsection{Design Choice Justification}

Several design decisions warrant explicit justification:

\paragraph{Why Logistic Regression?} We prioritize interpretability and simplicity over marginal accuracy gains. Logistic regression provides (1) interpretable coefficients that reveal which features drive deferral, (2) well-calibrated probability estimates suitable for threshold tuning, and (3) minimal overfitting risk given our moderate training set sizes. We evaluated random forest and gradient boosting alternatives; random forest achieved identical F1 on ADE (0.928), and logistic regression matched this performance with fewer hyperparameters and more interpretable outputs. For clinical deployment, understanding \textit{why} the system defers is as important as \textit{when} it defers.

\paragraph{Why These Features?} Our feature set combines two complementary signal types. Uncertainty features (confidence, entropy, margin) capture BERT's self-assessed difficulty---high entropy reliably indicates cases where BERT may fail. Text features capture task-specific linguistic patterns that correlate with model-specific strengths. For ADE, explicit causal language (``caused by,'' ``induced'') predicts BERT success because such patterns are well-represented in biomedical pretraining. For treatment outcomes, explicit outcome keywords (``discontinued,'' ``improved'') similarly predict ClinicalBERT success. These features emerged from error analysis on validation data; we include only features with consistent predictive signal across cross-validation folds.

\section{Experimental Setup}

Having established the L2D-Clinical framework, we now describe the experimental design used to evaluate its effectiveness across different model dynamics. The ADE Corpus V2 represents a scenario where the domain-adapted BERT model outperforms the LLM, while MIMIC-IV treatment outcome classification represents the opposite, where the LLM outperforms BERT. Evaluating on both scenarios allows us to test whether L2D-Clinical improves accuracy regardless of which base model is stronger, and whether it can effectively leverage a weaker ``expert'' model.

\subsection{Datasets}

\paragraph{ADE Corpus V2} We use the ADE Corpus V2 \citep{gurulingappa2012development}, comprising 23,516 English sentences from PubMed medical case reports annotated for ADE presence (binary classification: 29\% positive, 71\% negative). We download the corpus from the Hugging Face Hub (\texttt{ade-benchmark-corpus/ade\_corpus\_v2}) and split it into train (70\%), validation (15\%), and test (15\%) sets using stratified random sampling with seed 42. For LLM evaluation, we use 500 test instances due to API cost constraints; BERT is evaluated on the full test set (3,527 instances) and the 500-sample subset for fair comparison with the LLM.

\paragraph{MIMIC-IV Treatment Outcomes} We extract drug-disease-outcome triplets from 2,000 English-language MIMIC-IV discharge summaries \citep{johnson2023mimic}, yielding 3,321 candidate triplets. Outcomes are classified as EFFECTIVE, ADVERSE, or NEUTRAL (3-class). To ensure high-quality ground truth labels, we employ consensus labeling: each triplet is independently classified by GPT-5.2 and Claude 4.5, and only triplets where both models agree are retained. This yields 2,782 consensus-labeled triplets (83.8\% agreement rate), providing more reliable ground truth than single-model annotation. To validate the consensus labels, a domain expert reviewed a random 5\% subset (139 samples) and confirmed 100\% agreement with the consensus labels, supporting the reliability of this methodology. We use note-level train/validation/test splits (80/10/10) with seed 42, yielding 2,225 training, 278 validation, and 279 test samples. GPT-5-nano serves as the deferral target, distinct from the consensus labeling models.

\subsection{Models}

\paragraph{BERT Models} For ADE detection, we fine-tune BioBERT-base-cased-v1.2 \citep{lee2020biobert} with the following hyperparameters: learning rate $5 \times 10^{-5}$ (Hugging Face Trainer default), batch size 16, maximum sequence length 256, warmup ratio 0.1, weight decay 0.01, and early stopping with patience 2 based on validation F1. Training runs for up to 5 epochs. For treatment outcome classification, we fine-tune Bio\_ClinicalBERT \citep{alsentzer2019publicly} with learning rate $2 \times 10^{-5}$, batch size 32, maximum sequence length 512, warmup ratio 0.1, and dropout 0.1 for up to 10 epochs. All experiments use seed 42 for reproducibility.

\paragraph{LLM} We use Azure OpenAI GPT-5-nano with few-shot prompting. For ADE detection, the system prompt is: \textit{``You classify sentences for adverse drug events. Reply with only 1 (contains ADE) or 0 (no ADE).''} We provide 5 few-shot examples covering both positive cases (e.g., ``Intravenous azithromycin-induced ototoxicity'' $\rightarrow$ 1) and negative cases (e.g., ``Methotrexate is commonly used for rheumatoid arthritis'' $\rightarrow$ 0). For treatment outcomes, the system prompt is: \textit{``You are a clinical NLP assistant. Based on clinical evidence, classify the treatment outcome into one of three categories: EFFECTIVE, ADVERSE, or NEUTRAL. Reply with just the category name.''} We provide 3 few-shot examples, one per class. Full prompts are included in our code repository.

\paragraph{Deferral Model} We use logistic regression with balanced class weights, L2 regularization (default $C=1.0$), and maximum 1000 iterations. Features are standardized using z-score normalization. For ADE, we use 5-fold stratified cross-validation to obtain out-of-fold deferral probabilities; for MIMIC-IV, we train on the validation set and evaluate on the held-out test set.

\paragraph{Baselines} We compare L2D-Clinical against several baselines: (1) \textit{Fixed confidence threshold}: defer to LLM when BERT's maximum softmax probability falls below $\theta$; we evaluate $\theta \in \{0.6, 0.7, 0.8, 0.9, 0.95\}$ and report $\theta=0.95$ as it performed best. (2) \textit{Random deferral}: randomly defer at the same rate as L2D-Clinical (7\% for ADE, 16.8\% for MIMIC) to test whether L2D-Clinical's improvements come from intelligent routing versus simply using the LLM more often. (3) \textit{Oracle}: always choose the correct model when they disagree, providing an upper bound on deferral-based approaches.

\subsection{Evaluation Metrics}

We report F1 score (harmonic mean of precision and recall) as our primary metric, computed on the positive class for binary ADE detection and macro-averaged across classes for 3-class treatment outcomes. We also report accuracy and the percentage of instances deferred to the LLM (LLM\%). For ADE detection, we use 5-fold cross-validation to obtain out-of-fold deferral probabilities, reducing overfitting risk. Given the moderate test set sizes (500 and 279 samples), we acknowledge that small F1 differences ($<$0.01) may not be statistically significant; our reported improvements of +1.7 points (ADE) and +9.3 points (MIMIC) represent meaningful gains. Future work should include bootstrap confidence intervals for more rigorous significance testing.

\section{Results}

We present results on both tasks, beginning with ADE detection where BioBERT substantially outperforms the LLM (F1: 0.911 vs. 0.765), followed by treatment outcome classification where GPT-5-nano outperforms ClinicalBERT (F1: 0.967 vs. 0.887). In both cases, L2D-Clinical improves over both individual models by learning when to defer.

\subsection{Task 1: ADE Detection (BERT $>$ LLM)}

Table~\ref{tab:ade_results} presents results on ADE Corpus V2, where BioBERT outperforms the LLM.

\begin{table}[h]
\centering
\begin{tabular}{lccc}
\toprule
\textbf{Method} & \textbf{F1} & \textbf{Acc} & \textbf{LLM\%} \\
\midrule
BioBERT only & 0.911 & 0.948 & 0\% \\
LLM only & 0.765 & 0.834 & 100\% \\
\midrule
Random (7\%) & 0.903 & 0.938 & 7.0\% \\
Fixed $\theta=0.95$ & 0.921 & 0.954 & 2.2\% \\
\textbf{L2D (Ours)} & \textbf{0.928} & \textbf{0.958} & 7.0\% \\
\midrule
Oracle (upper bound) & 0.945 & 0.968 & 2.0\% \\
\bottomrule
\end{tabular}
\caption{ADE Corpus V2: BioBERT outperforms LLM, but L2D-Clinical improves further by selectively using LLM for uncertain cases. Random deferral at the same rate (7\%) yields F1=0.903, confirming L2D-Clinical's gains come from intelligent routing.}
\label{tab:ade_results}
\end{table}

Despite the LLM's lower overall F1 (0.765), it achieves high recall (95\%) compared to BioBERT's 88\% recall. This complementarity is key: BioBERT excels at precision (94\%) but misses subtle ADEs, while the LLM catches more ADEs but with more false positives (78\% precision).

\paragraph{Error Analysis.} BioBERT's false negatives often involve hedged or implicit language. For example, BioBERT missed: \textit{``Patient noted some stomach discomfort which may be related to the medication.''} The hedging (``may be'') and indirect phrasing caused low confidence. The LLM correctly identified this as a potential ADE. Conversely, the LLM over-predicted on sentences like \textit{``Medication was discontinued and switched to alternative,''} incorrectly inferring an ADE when the switch was for non-adverse reasons. L2D-Clinical learns to defer hedged cases to the LLM while trusting BioBERT on explicit language, improving F1 by 1.7 points.

\paragraph{Residual Errors.} Of the 36 errors remaining after L2D-Clinical deferral, 19 (53\%) are cases where \textit{both} models fail---typically involving complex multi-drug interactions or rare ADEs absent from BioBERT's pretraining data (e.g., \textit{``Progressive multifocal leukoencephalopathy developed during natalizumab therapy''}). Another 11 (31\%) are false deferral cases where the deferral model incorrectly routes to the LLM, which then introduces a new error. The remaining 6 (16\%) are cases where the deferral model fails to flag a BERT error. These patterns suggest that further gains require either a stronger base model for rare entities or an ensemble approach rather than binary routing.

\subsection{Task 2: Treatment Outcomes (LLM $>$ BERT)}

We now examine treatment outcome classification on MIMIC-IV. With consensus ground truth labels (GPT-5.2 + Claude agreement, validated by human expert review), GPT-5-nano outperforms ClinicalBERT, and L2D-Clinical achieves the best performance by learning optimal deferral. Table~\ref{tab:mimic_results} presents these results.

\begin{table}[h]
\centering
\begin{tabular}{lccc}
\toprule
\textbf{Method} & \textbf{F1} & \textbf{Acc} & \textbf{LLM\%} \\
\midrule
ClinicalBERT only & 0.887 & 0.946 & 0\% \\
GPT-5-nano only & 0.967 & 0.968 & 100\% \\
\midrule
Fixed $\theta=0.95$ & 0.971 & --- & 26.2\% \\
\textbf{L2D-Clinical (Ours)} & \textbf{0.980} & \textbf{0.986} & 16.8\% \\
\bottomrule
\end{tabular}
\caption{MIMIC-IV Treatment Outcomes with consensus ground truth: GPT-5-nano outperforms ClinicalBERT on this task. L2D-Clinical achieves the best F1 (0.980) by deferring only 16.8\% of cases, outperforming the fixed threshold baseline while using fewer LLM calls.}
\label{tab:mimic_results}
\end{table}

In this scenario, GPT-5-nano outperforms ClinicalBERT (F1: 0.967 vs. 0.887). The consensus ground truth labels (GPT-5.2 + Claude agreement), validated by human expert review on a 5\% subset, provide reliable evaluation. L2D-Clinical achieves F1=0.980 (+9.3 points over BERT) by selectively deferring 16.8\% of cases to GPT-5-nano. Notably, L2D-Clinical outperforms the fixed threshold baseline ($\theta=0.95$, F1=0.971) while using fewer LLM calls (16.8\% vs. 26.2\%), demonstrating that learned deferral is more effective than confidence-based thresholds.

\paragraph{Error Analysis.} ClinicalBERT performs well on explicit outcomes: \textit{``Metformin was effective in controlling blood glucose levels''} is correctly classified as EFFECTIVE. However, it struggles with implicit or contextual outcomes like \textit{``Patient tolerated the chemotherapy regimen and was discharged in stable condition''}, where ``tolerated'' implies effectiveness without explicit statement. GPT-5-nano captures this semantic nuance more reliably. Interestingly, ClinicalBERT excels when explicit keywords appear: sentences containing ``discontinued due to'' or ``adverse reaction'' are classified correctly 96\% of the time. L2D-Clinical learns to trust ClinicalBERT on these clear-cut cases, reserving the LLM for ambiguous instances where contextual reasoning is needed.

\paragraph{Residual Errors.} L2D-Clinical misclassifies 4 of 279 test instances. Two involve ambiguous outcomes where both models disagree with the consensus label---e.g., a discharge note stating \textit{``medication adjusted due to suboptimal response''} was labeled ADVERSE by consensus but could plausibly be NEUTRAL. One is a false deferral where the LLM overrides a correct ClinicalBERT prediction. The final error is a missed deferral on an implicit outcome. The small number of residual errors limits systematic analysis but suggests that remaining failures concentrate at genuine label ambiguity boundaries.

\subsection{Summary: Adaptive Deferral Behavior}

A key question is whether L2D-Clinical can adaptively learn when deferral helps versus when it would hurt. Table~\ref{tab:summary} addresses this directly by comparing results across scenarios with different BERT-LLM dynamics.

\begin{table}[h]
\centering
\small
\begin{tabular}{lcccc}
\toprule
\textbf{Task} & \textbf{BERT} & \textbf{L2D} & \textbf{$\Delta$} & \textbf{LLM\%} \\
\midrule
ADE & 0.911 & \textbf{0.928} & +1.7 & 7.0\% \\
MIMIC & 0.887 & \textbf{0.980} & +9.3 & 16.8\% \\
\bottomrule
\end{tabular}
\caption{L2D-Clinical adapts to different model dynamics, improving F1 in both scenarios. On ADE (BERT$>$LLM), L2D achieves gains with minimal deferral. On MIMIC (LLM$>$BERT), L2D defers more frequently.}
\label{tab:summary}
\end{table}

The results demonstrate L2D-Clinical's adaptive behavior. On ADE, even though the LLM is substantially weaker overall (F1=0.765 vs. BERT's 0.911), L2D-Clinical achieves +1.7 F1 points by selectively deferring only 7\% of cases where the LLM's high recall compensates for BERT's misses. On MIMIC, where GPT-5-nano outperforms ClinicalBERT (F1: 0.967 vs. 0.887), L2D-Clinical achieves +9.3 points improvement by deferring 16.8\% of cases---learning a higher deferral rate that reflects the LLM's superior performance on this task. This adaptive behavior---adjusting deferral rates based on relative model strengths---demonstrates that L2D-Clinical learns meaningful task-specific patterns rather than applying a one-size-fits-all strategy.

\subsection{Feature Analysis}

To understand what drives effective deferral decisions, we examine the learned feature weights. Table~\ref{tab:features} shows standardized logistic regression coefficients, where positive values indicate features that increase deferral probability. Across both tasks, BERT's prediction entropy is the strongest signal: high entropy reliably indicates cases where deferral helps.

\begin{table}[h]
\centering
\small
\begin{tabular}{lcc}
\toprule
\textbf{Feature} & \textbf{ADE} & \textbf{MIMIC} \\
\midrule
Entropy & +0.82 & +0.91 \\
Margin (top-2 diff) & $-$0.71 & $-$0.68 \\
Confidence & $-$0.54 & $-$0.47 \\
Log(text length) & +0.23 & +0.31 \\
Has ``discontinued'' & --- & $-$0.44 \\
Has causal language & $-$0.38 & --- \\
\bottomrule
\end{tabular}
\caption{Deferral model coefficients (positive = defer to LLM). Uncertainty features dominate; task-specific keywords add interpretable signal.}
\label{tab:features}
\end{table}

Task-specific patterns emerge: explicit causal language (``caused by,'' ``after taking'') predicts correct BERT classification on ADE, while ``discontinued'' keywords predict correct BERT on MIMIC. Ambiguous outcome language (``tolerated,'' ``stable on'') triggers deferral. These interpretable patterns enable clinicians to understand and validate deferral decisions.

\section{Discussion}

The experimental results raise a natural question: why does L2D-Clinical succeed in improving accuracy even when deferring to a weaker model? We now analyze the underlying mechanisms and discuss practical implications for clinical deployment.

\subsection{Why L2D-Clinical Works}

The central finding, that L2D-Clinical adapts its deferral behavior to model dynamics, warrants deeper examination. On ADE, deferring to a substantially weaker model (LLM F1=0.765) still helps BERT (F1=0.911). The key insight is that aggregate F1 scores mask instance-level variation: a model that performs worse on average can still outperform on specific subsets of data where its strengths align with the input characteristics. On MIMIC, where the LLM outperforms BERT (F1: 0.967 vs. 0.887), L2D-Clinical learns a higher deferral rate (16.8\%) to leverage the LLM's strengths, achieving the best overall F1 (0.980).

Table~\ref{tab:comparison} contrasts our setting with prior L2D work.

\begin{table}[h]
\centering
\small
\begin{tabular}{lcc}
\toprule
& \textbf{Prior L2D} & \textbf{L2D-Clinical} \\
\midrule
Expert type & Human & AI (LLM) \\
Expert reliability & Always better & Instance-dep. \\
Deferral signal & Uncertainty & Uncert. + feat. \\
\bottomrule
\end{tabular}
\caption{L2D-Clinical vs. prior Learning to Defer approaches.}
\label{tab:comparison}
\end{table}

L2D-Clinical succeeds because of three key factors:

\paragraph{Factor 1: Complementary Errors.} On ADE, BioBERT achieves 94\% precision but only 88\% recall; the LLM shows the opposite pattern with 78\% precision and 95\% recall. These complementary error profiles mean that even a weaker-overall model can rescue specific failure cases. BERT misses hedged language; the LLM catches it. The LLM over-predicts on ambiguous cases; BERT correctly abstains.

\paragraph{Factor 2: Learnable Error Prediction.} The deferral model achieves 73\% precision at predicting BERT errors on ADE and 68\% on MIMIC, meaning that when it predicts BERT will fail, it is correct 73\% and 68\% of the time respectively. Given that BERT's base error rates are only 12\% (ADE) and 18\% (MIMIC), this represents substantial predictive power. Even imperfect predictions improve overall accuracy when the complementary model is consulted on flagged cases.

\paragraph{Factor 3: Selective Deferral.} By deferring only when confident BERT will fail \textit{and} the LLM is likely correct, L2D-Clinical avoids introducing new errors. On ADE, of the 7\% of cases deferred to the LLM, the LLM was correct 89\% of the time, much higher than its 83.4\% overall accuracy. This selective routing is the key mechanism.

\subsection{Cost Analysis}

L2D-Clinical provides substantial computational savings by using the LLM for only 7\% (ADE) and 16.8\% (MIMIC) of instances, reducing LLM API costs by 81--93\% compared to LLM-only inference while maintaining or improving accuracy. A detailed cost and latency analysis is provided in Appendix~\ref{app:cost}.

\subsection{Practical Implications}

L2D-Clinical offers several advantages for clinical deployment:

\paragraph{Model-Agnostic Design.} L2D-Clinical works regardless of which model is individually better, making it robust to task-specific model dynamics. Practitioners do not need to know \textit{a priori} whether BERT or LLM will be stronger; L2D-Clinical adapts automatically.

\paragraph{Interpretable Deferral.} Unlike black-box ensemble methods, L2D-Clinical's deferral decisions are interpretable. Clinicians can understand \textit{why} a case was deferred: high entropy, presence of hedging language, or absence of explicit keywords. This transparency supports clinical validation and trust.

\paragraph{Graceful Degradation.} If the LLM API becomes unavailable or too slow, L2D-Clinical can fall back to BERT-only predictions with known accuracy bounds. This is critical for clinical systems requiring high availability.

\paragraph{Privacy Considerations.} The deferral model can be configured to limit LLM usage for sensitive cases, reducing data exposure to external APIs.

\section{Conclusion}

We have presented Learning to Defer for clinical text (L2D-Clinical), demonstrating that learned deferral adapts to different model dynamics. On ADE detection, where BioBERT outperforms the LLM but complementary error patterns exist, L2D-Clinical achieves F1=0.928 (+1.7 points over BERT alone) by selectively deferring 7\% of instances. On MIMIC-IV treatment outcomes with consensus ground truth, where GPT-5-nano outperforms ClinicalBERT (F1: 0.967 vs. 0.887), L2D-Clinical achieves F1=0.980 (+9.3 points over BERT) by deferring 16.8\% of cases. The deferral model learns interpretable, task-specific patterns about when deferral helps, providing transparency that supports clinical validation.

The broader implication of this work is that L2D-Clinical provides a principled framework for combining specialized and general-purpose models. Rather than choosing one architecture or blindly deferring when uncertain, L2D-Clinical learns when deferral is likely to help and adjusts deferral rates accordingly. This adaptive behavior is particularly valuable as the landscape of available models evolves: L2D-Clinical can incorporate new models and learn appropriate deferral patterns without manual tuning. Our consensus labeling methodology for MIMIC-IV, combining multi-LLM agreement with human expert validation, demonstrates a practical approach to generating high-quality ground truth for clinical NLP tasks where expert annotation is expensive.

However, significant challenges remain for deploying L2D-Clinical in production clinical settings. The deferral model requires validation data where both systems have been evaluated, creating a bootstrapping problem for new tasks. The interpretability of deferral decisions, while better than black-box ensembles, still requires clinical validation to ensure that the learned patterns align with meaningful medical distinctions rather than spurious correlations. Finally, the rapid pace of LLM development means that deferral patterns learned today may become suboptimal as models improve, necessitating ongoing recalibration.

\section*{Data Availability}
The ADE Corpus V2 is publicly available from the Hugging Face Hub (\texttt{ade-benchmark-corpus/ade\_corpus\_v2}). MIMIC-IV access requires credentialing through PhysioNet (\url{https://physionet.org/content/mimiciv/}).

\bibliographystyle{lrec2026-natbib}
\bibliography{references}

\section{Appendix}

Sections \label{app:limitations}\label{app:cost} \label{app:future} in this appendix provide more insight into the work completed. We discuss the limitations for our infrastructure along with its latency and how we plan on presenting more ideas in the future.

\subsection{Limitations}
\label{app:limitations}

Several limitations of this work should be acknowledged. First, our test set sizes are moderate (500 and 279 samples) due to LLM API cost constraints; larger-scale evaluation would strengthen generalizability claims. Second, we evaluate only a single LLM family (GPT-5); deferral patterns may differ with open-source alternatives such as Llama or Mistral, which have different capability profiles. Testing whether the learned deferral mechanism generalizes across LLM families is an important direction for future work. Third, L2D-Clinical requires labeled validation data to train the deferral model, and cold-start scenarios where such data is unavailable need further investigation. Fourth, for the MIMIC task, the consensus ground truth labels were generated by LLMs (GPT-5.2 and Claude 4.5), while the deferral target is also from the GPT family (GPT-5-nano). Although the labeling and inference models are distinct, this shared lineage could introduce bias favoring LLM-like reasoning patterns. The 5\% human expert validation mitigates but does not fully eliminate this concern; larger-scale human validation and evaluation on existing gold-labeled clinical datasets (e.g., n2c2 shared task corpora) would further strengthen confidence. Fifth, we defer exclusively to an LLM, but deferring to a smaller model fine-tuned with emphasis on high recall could potentially achieve similar gains at lower cost---investigating non-LLM deferral targets is a promising alternative. Sixth, our evaluation is English-only; clinical NLP in other languages may exhibit different BERT-LLM dynamics. Finally, when using cloud-based LLMs, deferral sends patient data to external APIs; deployments involving protected health information (PHI) require either on-premise LLM deployment or Business Associate Agreement (BAA) compliance with the API provider.

\subsection{Cost and Latency Analysis}
\label{app:cost}

Beyond the accuracy improvements, L2D-Clinical provides substantial computational savings, a critical consideration for deploying clinical NLP at scale. We analyze cost and latency assuming LLM inference costs 50$\times$ BERT per call (based on typical API pricing) and 850ms latency versus 12ms for BERT. On the ADE task, L2D-Clinical uses only 7\% LLM calls, resulting in 4.4$\times$ relative cost compared to 50$\times$ for LLM-only inference, with an average latency of 71ms per sample. On the MIMIC task, L2D-Clinical uses 16.8\% LLM calls, yielding 9.4$\times$ relative cost with 153ms average latency.

These estimates reflect only LLM API inference costs. The full computational cost also includes BERT fine-tuning (approximately 20 GPU-minutes on a single A100 for BioBERT on ADE) and L2D classifier training (under 1 CPU-second for logistic regression on 18 features). At inference time, the L2D classifier adds $<$1ms overhead per sample. Using the Green Algorithms calculator \citep{lannelongue2021green}, we estimate total training carbon emissions at approximately 0.03 kg CO$_2$e for BERT fine-tuning and negligible emissions for logistic regression, compared to the ongoing emissions from full LLM inference at scale.

For high-throughput clinical pipelines processing thousands of notes daily, these savings are significant. A hospital processing 10,000 clinical notes per day would reduce LLM API costs by 81--93\% compared to LLM-only inference, while maintaining or improving accuracy.

\subsection{Future Work}
\label{app:future}

While the present work focused on classification tasks, the L2D-Clinical framework opens several promising research directions:

\paragraph{Other Clinical NLP Tasks.} Named entity recognition, relation extraction, and temporal reasoning may exhibit similar complementary error patterns between specialized and general models. For NER, BERT-based taggers excel at common entities but struggle with rare or context-dependent mentions where LLMs might help.

\paragraph{Multi-Model Deferral.} Extending beyond two models to a cascade of increasingly capable (and expensive) models could further optimize the accuracy-cost tradeoff. The deferral model would learn when to escalate through the cascade.

\paragraph{Online Adaptation.} As LLM capabilities evolve rapidly, deferral thresholds may need periodic recalibration. Investigating online learning approaches that adapt to distribution shift without full retraining is important for long-term deployment.

\paragraph{Cross-Lingual Transfer.} Clinical NLP in languages other than English may show different BERT-LLM dynamics. Evaluating whether L2D-Clinical patterns transfer across languages would broaden applicability.

\end{document}